\pdfoutput=1
\documentclass{article} %
\usepackage{iclr2027_conference,times}

\usepackage{amsmath,amsfonts,bm}

\def\eqref#1{equation~\ref{#1}}
\def\1{\bm{1}}

\DeclareMathAlphabet{\mathsfit}{\encodingdefault}{\sfdefault}{m}{sl}
\SetMathAlphabet{\mathsfit}{bold}{\encodingdefault}{\sfdefault}{bx}{n}

\usepackage{hyperref}
\usepackage{url}
\usepackage{graphicx}
\usepackage{booktabs}
\usepackage[most]{tcolorbox}

\definecolor{tkframe}{HTML}{1D5A5E}
\definecolor{codelink}{HTML}{2A78D6}
\definecolor{tkback}{HTML}{F2F7F6}
\newtcolorbox{question}{enhanced, breakable=false, colback=tkback, colframe=tkframe, boxrule=0.9pt, arc=3.5mm,
  left=7pt, right=7pt, top=9pt, bottom=5pt, before skip=12pt, after skip=10pt,
  title={\textsc{Question}}, fonttitle=\normalsize, coltitle=white,
  attach boxed title to top left={xshift=5mm, yshift=-3.2mm},
  boxed title style={colback=tkframe, colframe=tkframe, arc=2mm, boxrule=0pt, left=5pt, right=5pt, top=1.5pt, bottom=1.5pt}}

\title{Spotter: Let the Embodied Model Lead, \\ and the VLM Reflect for It}

\author{Long Li$^{1*}$, Qichao Zhao$^{2*}$, Yue Yang$^{3}$, Fan Xu$^{4}$, Zhe Wang$^{1}$, \\
\textbf{Alan Wee-Chung Liew$^{1}$, Chao Qu$^{5}$, Heng Tao Shen$^{6}$, Shirui Pan$^{1\dagger}$} \\
{\normalfont\small $^{1}$Griffith University \quad $^{2}$IIIS, Tsinghua University \quad $^{3}$MainCode} \\
{\normalfont\small $^{4}$Tencent \quad $^{5}$Fudan University \quad $^{6}$Tongji University} \\
{\normalfont\small $^{*}$Equal contribution \quad $^{\dagger}$Corresponding author} \\
{\normalfont\small Email: Long Li \texttt{<long.li@griffithuni.edu.au>}}
}

\iclrfinalcopy %
\begin{document}

\maketitle
\lhead{Preprint}

\begin{abstract}
Current embodied models do not respond to their own failures, although what just went wrong could inform a small adjustment on the next attempt, the kind of reflection behind the gains of thinking in language models. We test whether they can repair a known error, which requires producing a correction and judging whether it is right. Stopped at a failure and allowed to retry, they seldom repair it through their own randomness or from a language description of the error, and best-of-N selection cannot pick the successful candidate after a failure. We attribute this to training only on successful demonstrations and to inputs too narrow to show what went wrong, and conclude that reflection must come from a vision-language model (VLM), which takes in far more information, such as the episode history and text, and is more general. Prior VLM-led work has the VLM plan every step and invoke the embodied model as a tool, placing the VLM on the critical path. We propose Spotter, which reverses the roles: the embodied model leads and executes continuously, while the VLM runs in parallel, monitors through a lightweight local screener, intervenes only when an error is detected, reflects on and corrects it, and returns control. We run Spotter with Qwen and with GPT as the VLM, and both improve the embodied models; with GPT, Spotter improves Cosmos Policy and $\pi_{0.5}$ by 5.6 and 7.5 percentage points on RoboCasa, and raises $\pi_{0.5}$ from 47.2\% to 57.0\% on the Hard setting of RoboTwin 2.0 and from 53\% to 83\% on a real robot. Because the VLM steps in only when an error is confirmed, a successful episode with Qwen takes only 13 to 16\,s longer than with the embodied model alone and about 70\% less time than with a VLM-led baseline using the same model. Our code is available at {\hypersetup{pdfborder={0 0 0}}\href{https://github.com/zqc3117/Spotter}{\textcolor{codelink}{github.com/zqc3117/Spotter}}}.
\end{abstract}

\section{Introduction}

\begin{figure}[t]
\begin{center}
\includegraphics[width=\linewidth]{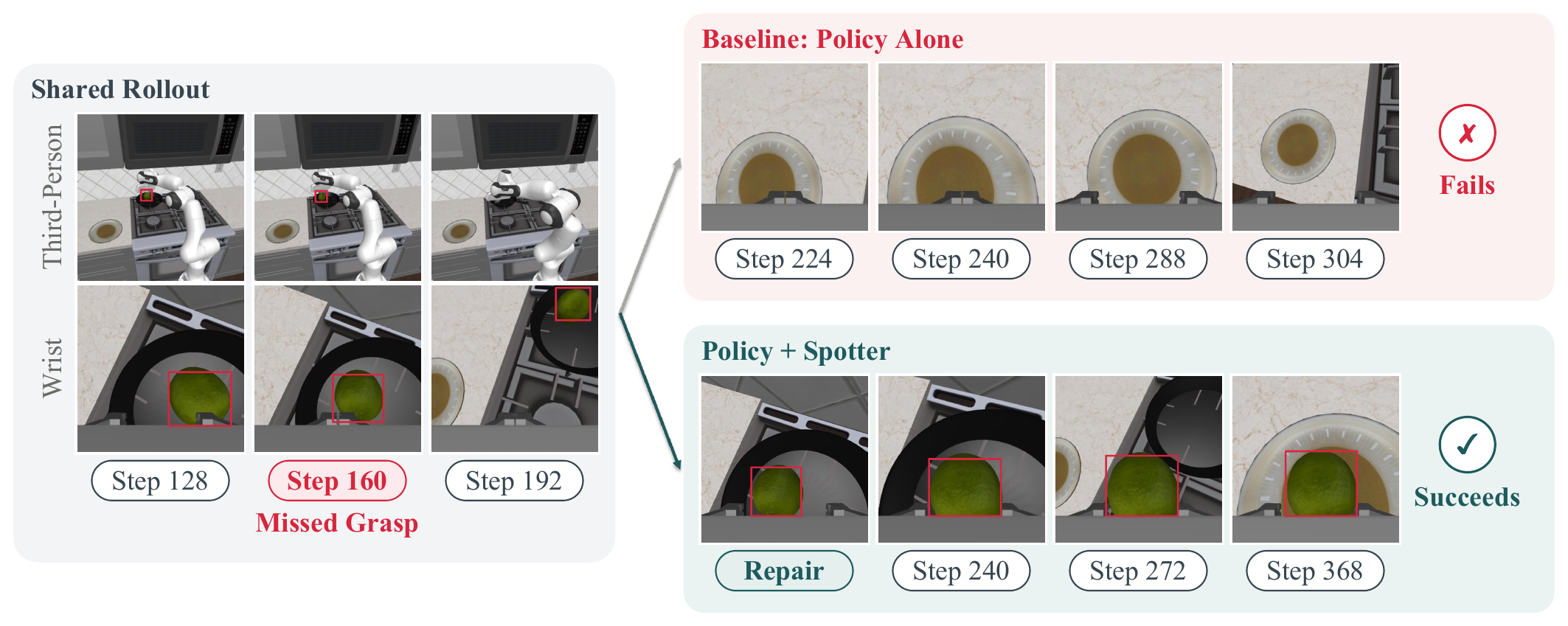}
\end{center}
\caption{The embodied model alone and with Spotter after the same missed grasp (task: move the lime from the pan to the plate). The grasp misses at step 160; by step 192 the arm heads for the plate. Alone (top), the model never notices, replaying the shape of a successful trajectory. With Spotter (bottom), the VLM notices the miss only after the arm has moved on: by step 192 it has already left the pan. This delay can be undone: the repair returns the arm to its pose before the miss and, since the failed attempt was off to the left, shifts the gripper right to face the lime.}
\label{fig:failure}
\end{figure}

Embodied models have advanced rapidly, from generalist policies that map images and language to actions \citep{pmlr-v229-zitkovich23a,pmlr-v270-kim25c} to models built on vision-language backbones or pretrained video models that succeed on standard manipulation benchmarks \citep{BlackK-RSS-25,pmlr-v305-black25a,bjorck2025gr00t,geminiroboticsteam2025geminiroboticsbringingai,kim2026cosmos}. Yet they cannot correct their own errors. In Figure~\ref{fig:failure} the grasp misses the lime, and the policy carries nothing toward the plate as if the object were in hand. Independent evaluations agree: failures are dominated by missed grasps and premature releases \citep{zhang2025experiences}, policies rarely recover from them \citep{yu2026benchmarking}, and they keep executing actions that no longer solve the task \citep{Zhao_2026_CVPR,pmlr-v270-agia25a}. In manipulation this is costly, because whether a grasp holds or a placement is stable is known only after the fact \citep{he2026fawam}, so errors must be caught and corrected during execution.

Correcting an error takes two abilities, producing a correction and judging whether it is right; current embodied models have neither (Section~\ref{sec:eval}). Stopped at a failure and given fresh attempts, they repair fewer than one in ten, whether they resample, are told what went wrong, or are fine-tuned on recovery data; and test-time selection, which samples several candidates and lets a scorer keep one, fails because after a failure the scorer cannot tell which will succeed. Two things are missing. The policy and its scorers learn from successful demonstrations, in which failures never appear \citep{lin2025failsafe,11575083}, so naive retries repeat the original error \citep{hao2026far}, and fine-tuning does not help because recovery data are scarce in the policy's own rollouts and cover few kinds of failure. And they see only the current moment, whereas telling a recovery from a repeated failure needs the history of the episode: even a vision-language model (VLM) shown only the current images judges no better than random. Given that history, a VLM can supply both: it reads earlier attempts and text, and generalizes beyond the demonstrations, routinely correcting its own errors \citep{NEURIPS2023_1b44b878,madaan2023self}; Section~\ref{sec:analysis} shows that it intervenes where it should and repairs errors the embodied model does not.

Prior work that pairs the two is VLM-led \citep{10160591,zhang2026harness,wang2026towards}: the VLM plans every step and calls the embodied model as a tool. With the embodied model as a fast System~1 and the VLM as a slow System~2 \citep{bjorck2025gr00t}, control switches wholesale: while the VLM deliberates, the embodied model waits, and while it acts, nothing watches. The VLM must also fix in advance how long each stretch runs: short stretches waste time on checks the policy does not need, and long ones let it go far down a failure.

We propose Spotter, which reverses the roles, much as a person lets a skilled action run and calls in deliberate thought only when something seems off. The embodied model leads and executes continuously. A lightweight local screener screens every chunk and forwards suspicious windows to the VLM, which runs in parallel; how often the VLM looks thus depends on events rather than a fixed schedule, and execution pauses when an error is confirmed or the embodied model gets $K$ chunks ahead of the supervisor. The VLM then repairs the error with a short plan over motion primitives and reflects on it: it states what it expects to see and returns control, then checks the next supervised window and, if the expectation failed, tries a different kind of repair. Our contributions are threefold:\\
(1) an evaluation of the error-correction ability of current embodied models, testing whether they can produce a correction and judge whether it is right;\\
(2) Spotter, an embodied-led framework in which a VLM supervises in parallel and intervenes only on failure, without training the policy;\\
(3) experiments with Cosmos Policy and $\pi_{0.5}$ on RoboCasa and $\pi_{0.5}$ on the Hard setting of RoboTwin 2.0 and a real robot, with Qwen or GPT as the VLM, showing that Spotter raises success rates in far less time than a VLM-led design, with successful episodes taking time of the same order as with the embodied model alone.

\section{Related Work}
\label{sec:related}

\textbf{Robot Agents.} Pairing a language model with an embodied model is an emerging paradigm for open-vocabulary manipulation, and existing systems are VLM-led, with the language model deciding. It may write programs over perception and motion APIs \citep{10160591}, as benchmarked by CaP-X \citep{fu2026capx}; issue a language command every second, or whenever the user speaks, to an embodied model fine-tuned to follow such commands, as in Hi Robot \citep{pmlr-v267-shi25d}, which targets open-ended instructions rather than failures; invoke a frozen embodied model as one primitive among several, as in Harness VLA \citep{zhang2026harness}; or run an evaluator after every action to diagnose failures, as in Towards the Harness of Embodied Agents \citep{wang2026towards}. RACER \citep{11127799} queries a VLM supervisor at every step for a language hint, training both supervisor and actor on failure-recovery data. In all of these, execution waits on the language model at every decision, so latency grows with task length rather than with the number of failures. Spotter runs the VLM in parallel with a leading embodied model, pauses execution on a confirmed error or at the bounded lead limit, and lets the VLM repair the failure itself with a short primitive plan (Table~\ref{tab:comparison}).

\begin{table}[t]
\caption{Comparison with representative prior work. VLM-led designs wait on the external model; Spotter runs both in parallel. \emph{Semantic}: corrects errors invisible in telemetry, such as reaching for the wrong object. \emph{Training}: needs training beyond the off-the-shelf embodied model.}
\label{tab:comparison}
\begin{center}
\begin{tabular*}{\linewidth}{@{\extracolsep{\fill}}lllcc@{}}
\toprule
\textbf{Method} & \textbf{Architecture} & \textbf{Repair} & \textbf{Semantic} & \textbf{Training} \\
\midrule
Hi Robot            & VLM-led  & none            & --      & Yes        \\
Harness VLA         & VLM-led  & failure memory  & Partial & No         \\
Towards the Harness & VLM-led  & diagnose, retry & Yes     & No         \\
RACER               & VLM-led  & language hint   & Yes     & Yes        \\
\midrule
\textbf{Spotter (ours)} & \textbf{Parallel} & \textbf{primitive plan} & \textbf{Yes} & \textbf{No} \\
\bottomrule
\end{tabular*}
\end{center}
\end{table}

\textbf{Failure Explanation and Recovery.} Language models have long been used to explain and recover from robot failures. Inner Monologue \citep{pmlr-v205-huang23c} feeds success detection and scene descriptions back to a language-model planner, which replans in language, and DoReMi \citep{10802284} has the planner also write constraints that a VLM checks continuously during execution, triggering recovery when one is violated. Both are VLM-led: the language model plans over a fixed set of skills, and recovery means replanning at the skill level. Another line learns from failures: RoboFAC \citep{ye2025robofac} builds a failure-centric dataset of erroneous trajectories and question-answer pairs and fine-tunes a multimodal model that acts as an external supervisor of a vision-language-action model (VLA), and AIC-MLLM \citep{pmlr-v270-xiong25a} fine-tunes a multimodal model to correct its own SE(3) contact poses on articulated objects from interaction feedback. Spotter needs neither failure data nor training: the embodied model leads, a VLM checks it in parallel, and the VLM repairs an error itself with motion primitives rather than handing a plan or instruction back to it.

\textbf{Test-Time Scaling.} Spending extra inference compute is standard in language modeling, through beam search \citep{NEURIPS2023_81fde95c}, majority voting \citep{wang2023selfconsistency}, and verifier-based best-of-N selection \citep{cobbe2021training}. The recipe transfers to embodied models: a stochastic policy can be sampled repeatedly at one state, and a model that also predicts future observations attaches an imagined outcome to each candidate. Candidates are ranked by a learned value, as in Cosmos Policy \citep{kim2026cosmos}; by a separately trained vision-language progress critic, VLAC \citep{zhang2026a}; or without training, by Consistency-Consensus \citep{ruan2026future} (with its non-deployable upper bound Consistency-Exploring), GeoBoN \citep{zhao2026testtimescalingworldaction}, and RCS from $\tau_0$-WM \citep{zhou2026tau_0} (definitions in Appendix~\ref{app:eval}). All assume that the scorer can tell good candidates from bad ones; since policy and scorers alike learn from successful demonstrations, we test this right after a failure, asking which candidate repairs it (Section~\ref{sec:eval}), and find that the assumption does not hold.

\section{Method}
\label{sec:method}

\subsection{Error-Correction Ability of Current Embodied Models}
\label{sec:eval}

We first evaluate the error-correction ability of current embodied models: an oracle marks when an error has occurred and interrupts execution there, and we measure how well the model then corrects it. Correction involves two abilities, producing a corrected action and judging whether a correction is right. We pose each as a question and answer it with a dedicated experiment using the world action model Cosmos Policy and the VLA $\pi_{0.5}$ on the pick-and-place tasks of RoboCasa, where an oracle tells unambiguously whether the object has been grasped.

\begin{question}
\textbf{Q1: Once an error is caught by the oracle, can the embodied model produce a correction?}
\end{question}

We interrupt execution at each failure and give the model three attempts to correct it under each of three protocols, counting the failure as corrected if any attempt lifts the object (Appendix~\ref{app:eval}).

\textbf{Resampling.} We restore the simulator to one chunk before the failed grasp and let the model act again under three random seeds, executing all three continuations; with no signal that anything went wrong, this tests whether its own randomness alone can repair the error.

\textbf{Language feedback.} Execution stops at the failed grasp without rewinding; a VLM shown the current camera views is told that the grasp has failed and asked what went wrong and what to do differently, and its answer becomes the model's instruction for another attempt, up to three rounds.

\textbf{Recovery fine-tuning.} Following \citet{11575083,hao2026far,Zhao_2026_CVPR}, we fine-tune the policy on recovery data collected from its own rollouts and repeat the resampling protocol; in our runs only roughly 3\% of ordinary rollouts yield recovery data meeting our requirements, so the data are scarce and cover few of the many kinds of failure.

\begin{figure}[!ht]
\begin{center}
\includegraphics[width=\linewidth]{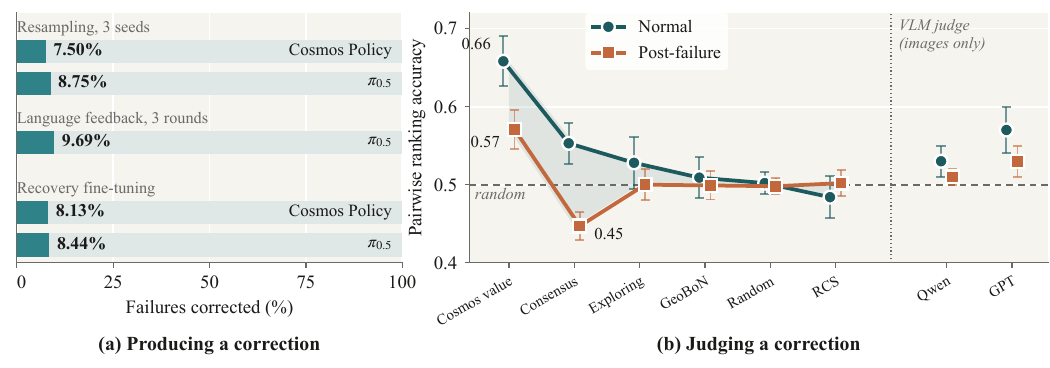}
\end{center}
\caption{Error-correction ability of current embodied models. (a) Fraction of failed grasps that the model corrects when given three more attempts, before and after fine-tuning on recovery data, over 320 failed grasps per setting. (b) Pairwise ranking accuracy of test-time scorers and of the VLM judge given only images around the saved state, at normal states and after a failed grasp; error bars are 95\% bootstrap confidence intervals.}
\label{fig:eval}
\end{figure}

Figure~\ref{fig:eval}(a) reports the outcome. None of the three works: fewer than one failure in ten is corrected, even though execution is interrupted right at the failure and three attempts are allowed. Language feedback is only slightly better than resampling: being told what went wrong leaves the model about as likely to succeed as drawing a new seed, because current embodied models do not take in an instruction and revise their behavior the way a language model does, and Cosmos Policy does not even support instructions other than the official ones it was trained with. Fine-tuning on recovery data does not help either (8.13\% for Cosmos Policy, 8.44\% for $\pi_{0.5}$), and in our runs the overall success rate stays close to the baseline.

\begin{question}
\textbf{Q2: Among several candidate corrections, can the successful one be identified?}
\end{question}

This is precisely what recent test-time scaling methods for embodied models set out to do: sample several candidates from the embodied model at the same state and let a scorer keep the best one. We find that, after a failure, none of them can reliably pick the successful candidate.

\textbf{Protocol.} Using Cosmos Policy, we save the simulator at a chosen state, restart the policy from that save eight times with different random seeds, and run each candidate for 32 steps, labeling it by whether the object ends up grasped. Within a state we pair every successful candidate with every failed one and report the fraction of pairs in which the scorer ranks the successful one higher, so 0.5 is random. We evaluate the value head of Cosmos Policy, Consistency-Consensus, Consistency-Exploring, GeoBoN and RCS, together with a random baseline that should land at 0.5. As a reference from outside the embodied model, we also score the same candidates with the VLM judge of Spotter (Section~\ref{sec:spotter}), Qwen or GPT-6 Astra, given only images around the saved state (Appendix~\ref{app:eval}).

\textbf{Post-failure and normal states.} In the setting the scorers would be used in, the save is taken the instant a grasp has failed, so the candidates are attempts to recover, some of which succeed. As a control showing what the scorers can do when conditions are in their favor, the save is taken one chunk before a grasp the policy completed successfully, on an ordinary trajectory, where some candidates grasp the object as usual and others miss.

Figure~\ref{fig:eval}(b) reports the outcome: at normal states the scorers have some skill, and after a failure they lose it. GeoBoN and RCS hover around random in both settings, while the value head of Cosmos Policy and Consistency-Consensus, clearly above random at normal states, show a large gap between the two settings, dropping sharply once the policy has failed.

Both results trace back to the embodied model: learned from successful demonstrations, which lack the states that follow a failure, it rarely produces a correction there, and recovery data collected from it are too scarce to supply one. Judging a correction, however, is hard even for a model that did not learn from these demonstrations when it sees only the current state: given the same images, the VLM judge orders success above failure 53\% of the time after a failure with GPT-6 Astra and 51\% with Qwen, against 57\% and 53\% at normal states. Candidates from the same state differ only slightly, and a single view does not show what has already been tried. Telling a recovery from a repeated failure takes the history of the episode: the failed attempt, what came before it, and text such as the robot's telemetry and the actions so far. We therefore bring in a vision-language model that reads a much longer stretch of the episode, including the failed attempt, together with this text, and that generalizes well enough to correct the embodied model's errors: instead of instructing the embodied model or choosing among its samples, it composes a repair plan over motion primitives and checks in the next supervised window whether it worked (Section~\ref{sec:spotter}); Section~\ref{sec:analysis} measures how often it corrects an error.

\begin{figure}[t]
\begin{center}
\includegraphics[width=\linewidth]{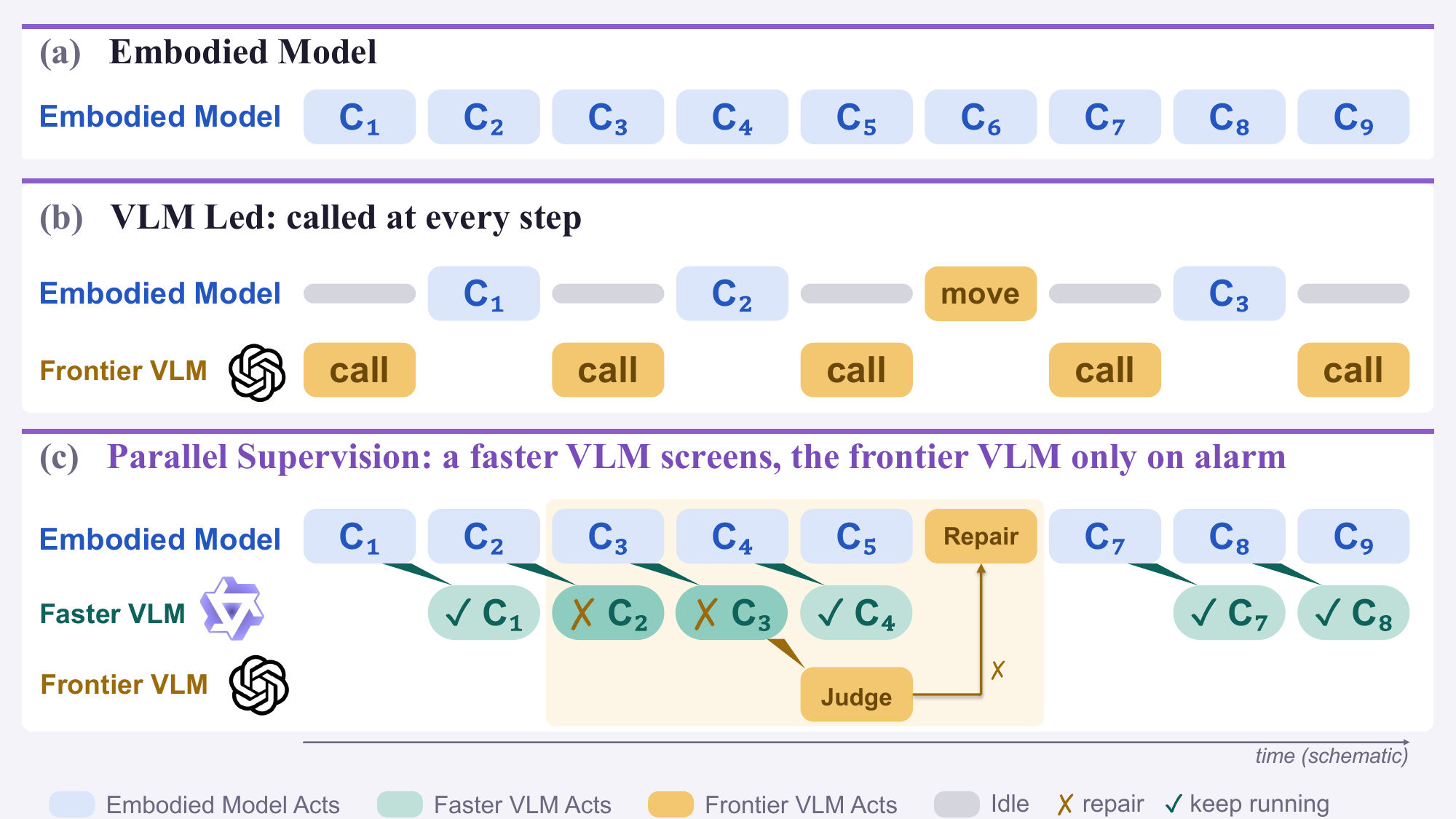}
\end{center}
\caption{Three ways to combine an embodied model with a VLM over time (block widths are schematic). (a) The embodied model acts alone. (b) VLM-led: the frontier VLM is called before every step, and the embodied model idles while it waits. (c) Spotter: a faster VLM screens each chunk in parallel, one chunk behind execution, and, on the routine path illustrated here, the frontier VLM is called after two consecutive flags. Once $C_2$ and $C_3$ are both flagged, the judge deliberates while the embodied model executes $C_5$; it is interrupted only when the error is confirmed, at $C_6$, for the repair, then resumes. For clarity each drawn check covers one chunk and the lead limit $K$ is not drawn; in practice a delayed check takes in every completed chunk not yet checked, so none is discarded.}
\label{fig:framework}
\end{figure}

\subsection{Spotter}
\label{sec:spotter}

\textbf{Problem definition.} An embodied model $\pi$ performs a manipulation task described by a language instruction $\ell$: at each step $t$ it receives RGB images $I_t$ and the robot state $q_t$ (end-effector pose and gripper state), and outputs the next $H$ actions, $a_{t:t+H} = \pi(I_t, q_t, \ell)$. The task succeeds when a condition $\mathcal{G}$, such as ``the object is on the plate'', becomes true; the episode ends then, or else at the step limit $T$ set for the task. We keep $\pi$ frozen and add a supervisor that sees the same images and robot state and is never told whether $\mathcal{G}$ holds. The depth maps $D_t$ of the same cameras are used only by the executor to lift the pixel targets of a repair to 3-D; the VLM itself never sees them. The goal is to raise the success rate of $\pi$ within the same budget of $T$ steps, while overlapping supervision with policy execution.

\textbf{Parallel supervision.} Figure~\ref{fig:framework} compares three designs. Run alone (a), $\pi$ fails unnoticed; in VLM-led designs (b) such as Harness VLA \citep{zhang2026harness}, the VLM plans every step, so its latency is paid throughout the episode. Spotter (c) runs $\pi$ (System~1) and a VLM supervisor in parallel, with continuous attention between them. Most actions, such as moving the arm to a location, are routine and handled well by $\pi$, so the supervisor only has to catch the few that go wrong: a fast, high-recall \emph{screener} checks every chunk, and a frontier VLM (System~2), the \emph{judge}, examines flagged windows and verifies the outcome of a repair. The VLM receives only what the robot's own sensors provide at deployment: the camera images and robot state that $\pi$ sees and the actions $\pi$ has commanded, never privileged simulator state such as object poses or the success signal. The embodied model keeps running during supervision until an error is confirmed or the bounded lead limit is reached.

\textbf{Screening, repair, and reflection.} Let $C_i$ be the $i$-th chunk executed by $\pi$ and $z_i$ its record, the camera views and telemetry of the chunk. When $\pi$ finishes $C_i$ it moves straight on, and $z_i$ goes to the screener. With $j$ the latest chunk the supervisor has checked, $\pi$ may start chunk $i$ only if
\begin{equation}
i - j \le K,
\label{eq:lead}
\end{equation}
that is, it runs at most $K$ chunks ahead of the check; at this lead limit it stops at the chunk boundary until the check catches up. Apart from a confirmed error, this is the only time $\pi$ pauses, and it bounds staleness: when the judge reaches a verdict, the robot has moved at most $K$ chunks beyond what the judge saw. Because the screener can run slower than $\pi$, chunks may queue up, so each check takes in all chunks finished since the previous one, $z_{j+1:i}$, not only the latest. The screener $S$, a locally deployed Qwen model, gives a risk score and flags the chunks when the score reaches a threshold $\tau$:
\begin{equation}
r_i = S(z_{j+1:i}), \qquad f_i = \mathbf{1}[\, r_i \ge \tau \;\lor\; E_i \;\lor\; F_i \,].
\label{eq:screen}
\end{equation}
Here $E_i$ denotes empty-gripper evidence and $F_i$ a failed or invalid screening call (Appendix~\ref{app:setup}). We tune it to favor recall: $\tau$ is low, because a miss costs far more than a false alarm, and any screener failure counts as a flag. To limit false alarms, the judge $J$ is called after two consecutive screening checks are flagged, or immediately for empty-gripper evidence under the rule in Appendix~\ref{app:setup}. The next supervised window after a repair also receives an observation-only judge check. In a decision call the judge returns
\begin{equation}
(v, \delta, e, p) = J(z_{j+1:i}, h_i), \qquad v \in \{\emph{ok}, \emph{intervene}, \emph{giveup}\},
\label{eq:judge}
\end{equation}
where $h_i$ is the episode history, $v$ the verdict, $\delta$ a short free-text diagnosis, and, when it intervenes, $e$ an expectation and $p = (c_1, \dots, c_n)$, $c_k \in \mathcal{P}$, a plan of primitives (Table~\ref{tab:primitives}). If nothing is wrong, it returns \emph{ok} and $\pi$ continues undisturbed. Otherwise the plan runs segment by segment, and the executor lifts each image target, a pixel in a named camera view, to 3-D with its depth in $D_t$ (Appendix~\ref{app:primitives}). Unless the episode has ended, after each segment the judge sees a new image and the execution log and decides to continue, declare the problem fixed, or give up; $\pi$ then resumes from the repaired state, with no rollback. Budgets cap the segments per intervention and the interventions per episode, and every motion counts toward $T$. Every intervention also states an expectation, an observable fact for the next supervised window such as ``the gripper stays open wider than 20\,mm while the arm lifts''. In that verification window the judge may only observe, and must first say whether the expectation held; if it did, it must not repeat the repair, and if not, its next intervention must try a different kind of repair rather than a new offset. Without this step, the judge tends to repair and let go, unaware of whether it worked.

\begin{table}[t]
\caption{Primitives available to the judge.}
\label{tab:primitives}
\begin{center}
\begin{tabular*}{\linewidth}{@{\extracolsep{\fill}}ll@{}}
\toprule
\textbf{Primitive} & \textbf{Effect} \\
\midrule
Move to point      & Move the end effector to a back-projected target \\
Move by offset     & Translate the end effector by a relative displacement \\
Lift               & Raise the end effector \\
Rotate wrist       & Rotate the wrist by a given angle \\
Set gripper        & Open or close the gripper \\
Retreat            & Return the arm to a recorded earlier pose of the judge's choice \\
Move both arms     & Move the two arms synchronously \\
Act                & Run the supervised embodied model $\pi$ for a given chunk size \\
Hand back          & End the takeover and return control to the embodied model \\
\bottomrule
\end{tabular*}
\end{center}
\end{table}

\textbf{Issues raised by asynchrony.} The judge examines what the model produced up to $K$ chunks ago (Eq.~(\ref{eq:lead})); does the lag make a difference? Three cases arise. (i) \emph{No damage}: the faulty action has not changed the environment; it is passed to the VLM, which uses it to return the arm to its earlier position, undoing the action. (ii) \emph{Repairable damage}: an object is knocked over but still on the table, or pushed out of its original region on contact; the VLM can locate it again and complete the task, just as it could on a real robot. (iii) \emph{Irreversible damage}: the task cannot continue, for example because the object falls to the floor out of view. This is not a limitation of our framework: if an action ruins the environment the first time it is executed, paradigms (a) and (b) cannot recover either; such cases call for a more advanced approach, possibly a world model simulating the future.

\textbf{Physical reset and learning.} In the first case above, Spotter returns the arm, not the scene: Spotter records the arm pose at each chunk boundary, and to redo an action the judge uses the Retreat primitive to return to the recorded takeover pose or the pre-grasp anchor, without rolling back the simulator, so the whole method runs on a real robot. The simulator's reset is used only while learning a library of general lessons, free of privileged or task-specific information (Appendix~\ref{app:setup}); at test time the lessons are only retrieved, and nothing is reset.

\section{Experiments}
\label{sec:experiments}

We evaluate Spotter without an in-context demonstration (\textbf{zero-shot}, all table rows other than \emph{1-shot}) and with a single learned demonstration (\textbf{1-shot}). In the RoboCasa implementation, both settings retrieve from the same frozen library of general lessons; 1-shot additionally retrieves one demonstration from separate learning episodes (Section~\ref{sec:spotter}). Neither resets at test time.

\subsection{Experimental Setup}
\label{sec:setup}

\textbf{Benchmarks and embodied models.} We run three sets of experiments. (i) On RoboCasa \citep{Nasiriany-RSS-24}, a simulated benchmark and our main testbed, we use the released checkpoints of Cosmos Policy \citep{kim2026cosmos} and $\pi_{0.5}$ \citep{pmlr-v305-black25a} on all 24 official tasks (8 pick-and-place tasks and 16 that operate a mechanism such as a door, a drawer, or a knob), with 50 episodes each (1{,}200 episodes in total). (ii) On the Hard setting of RoboTwin 2.0 \citep{chen2026robotwin}, a simulated bimanual benchmark that tests transfer to a different simulator and embodiment, we use $\pi_{0.5}$ trained on clean data only, on 50 tasks with 10 episodes each. (iii) On a real Franka Research 3 robot, we use $\pi_{0.5}$ fine-tuned on teleoperated demonstrations for three tasks (Section~\ref{sec:real}). Episode seeds are fixed in advance. Qwen3.8-27B \citep{qwen38}, served locally, is the screener, and either Qwen itself or GPT-6 Astra is the judge, denoted \textbf{Spotter (Qwen)} and \textbf{Spotter (GPT)}. Every episode is run with and without Spotter under the same scene, instruction, and step budget, raised equally for a matched comparison; repair actions consume the same episode budget as policy actions. The supervisor learns from episodes disjoint from the test episodes and never sees the success signal. Task lists, seeds, hyperparameters, the settings of Harness VLA (run with Qwen), and the sources of Figure~\ref{fig:results} are in Appendix~\ref{app:setup}. Table~\ref{tab:context} lists the three Spotter configurations used in the tables.

\begin{table}[!htbp]
\caption{The three Spotter configurations. Telemetry: recent chunks shown as rows; older chunks are summarized. Turns: recent Qwen judge turns kept in full, in addition to the initial briefing; older turns are compacted. Context restart: the token threshold for restarting with a text recap. GPT uses server-side conversation continuation.}
\label{tab:context}
\begin{center}
\small
\begin{tabular*}{\linewidth}{@{\extracolsep{\fill}}llll@{}}
\toprule
 & \textbf{Short context} & \textbf{Full context} & \textbf{1-shot} \\
\midrule
Telemetry & last 3 & last 40 (Cosmos) / 10 ($\pi_{0.5}$) & as full \\
Turns & last 3 & last 20 & as full \\
Context restart & 22k tokens & 110k tokens & as full \\
Interventions per episode & 3 & 8 & 8 \\
Judge called after & 1 flag & 2 flags & 2 flags \\
Example & none & none & 1 episode \\
\bottomrule
\end{tabular*}
\end{center}
\end{table}

\subsection{Main Results}
\label{sec:main-results}

\begin{table}[!htbp]
\caption{Success rates (\%) on RoboCasa, 24 tasks with 50 episodes each (PnP: the 8 pick-and-place tasks; Non-PnP: the 16 tasks that operate a mechanism). \emph{Fixed retry}: on a flagged failure the arm retreats and control returns to the embodied model, with no judge. $\Delta$: gain over the embodied model at $1.8\times$ the step limit, the budget of Spotter. Harness VLA uses 40 rounds and 5,000 steps (Appendix~\ref{app:setup}); with \emph{privileged info} its VLM can also see the task's success predicate, which Spotter never receives. Best per column in bold.}
\label{tab:main}
\begin{center}
\small
\setlength{\tabcolsep}{1.2pt}
\newcommand{\up}[1]{\textcolor[HTML]{2E7D32}{+#1}}
\newcommand{\down}[1]{\textcolor[HTML]{C62828}{$-$#1}}
\begin{tabular*}{\linewidth}{@{\extracolsep{\fill}}lcccccccccccc@{}}
\toprule
 & \multicolumn{6}{c}{\textbf{Cosmos Policy}} & \multicolumn{6}{c}{$\boldsymbol{\pi_{0.5}}$} \\
\cmidrule(lr){2-7} \cmidrule(lr){8-13}
\textbf{Method} & \textbf{PnP} & $\Delta$ & \textbf{Non-PnP} & $\Delta$ & \textbf{All} & $\Delta$ & \textbf{PnP} & $\Delta$ & \textbf{Non-PnP} & $\Delta$ & \textbf{All} & $\Delta$ \\
\midrule
Embodied model & & & & & & & & & & & & \\
\quad -$1.0\times$ step limit & 53.2 & & 73.4 & & 66.7 & & 56.8 & & 65.4 & & 62.5 & \\
\quad -Screener + fixed retry & 52.2 & & 74.2 & & 66.9 & & 56.2 & & 65.0 & & 62.1 & \\
\quad -Oracle + fixed retry   & 53.5 & & 73.9 & & 67.1 & & 56.0 & & 66.6 & & 63.1 & \\
\quad -$1.8\times$ step limit & 55.2 & & 74.2 & & 67.9 & & 58.0 & & 67.5 & & 64.3 & \\
\midrule
Harness VLA (Qwen) & & & & & & & & & & & & \\
\quad -privileged info    & 57.5 & \up{2.3} & \textbf{78.2} & \up{4.0} & 71.3 & \up{3.4} & 62.2 & \up{4.2} & 72.5 & \up{5.0} & 69.1 & \up{4.8} \\
\quad -no privileged info & 57.5 & \up{2.3} & 69.0 & \down{5.2} & 65.2 & \down{2.7} & 60.0 & \up{2.0} & 60.8 & \down{6.7} & 60.5 & \down{3.8} \\
\midrule
Spotter (Qwen) & & & & & & & & & & & & \\
\quad -short context & 59.0 & \up{3.8} & 74.5 & \up{0.3} & 69.3 & \up{1.4} & 63.5 & \up{5.5} & 68.9 & \up{1.4} & 67.1 & \up{2.8} \\
\quad -full context  & 59.3 & \up{4.1} & 77.9 & \up{3.7} & 71.7 & \up{3.8} & 63.2 & \up{5.2} & 71.0 & \up{3.5} & 68.4 & \up{4.1} \\
\quad -1-shot        & 63.5 & \up{8.3} & 76.0 & \up{1.8} & 71.8 & \up{3.9} & 65.2 & \up{7.2} & 73.0 & \up{5.5} & 70.4 & \up{6.1} \\
\midrule
Spotter (GPT) & & & & & & & & & & & & \\
\quad -with screener & 58.8 & \up{3.6} & 75.6 & \up{1.4} & 70.0 & \up{2.1} & 65.0 & \up{7.0} & 70.8 & \up{3.3} & 68.8 & \up{4.5} \\
\quad -no screener   & 61.3 & \up{6.1} & 77.4 & \up{3.2} & 72.0 & \up{4.1} & \textbf{67.8} & \up{9.8} & 72.2 & \up{4.7} & 70.8 & \up{6.5} \\
\quad -1-shot        & \textbf{65.5} & \up{10.3} & 77.5 & \up{3.3} & \textbf{73.5} & \up{5.6} & 67.0 & \up{9.0} & \textbf{74.2} & \up{6.7} & \textbf{71.8} & \up{7.5} \\
\bottomrule
\end{tabular*}
\end{center}
\end{table}

\begin{figure}[!htbp]
\begin{center}
\includegraphics[width=\linewidth]{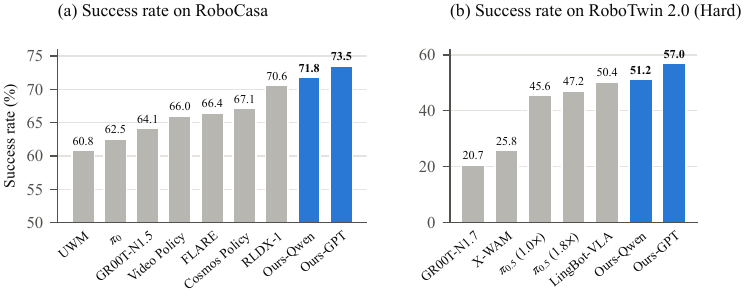}
\end{center}
\caption{Success rate compared with other embodied models on (a) RoboCasa and (b) the Hard setting of RoboTwin 2.0. The compared methods are described in Appendix~\ref{app:setup}.}
\label{fig:results}
\end{figure}

Table~\ref{tab:main} reports success rates on RoboCasa. A longer budget alone helps little, adding 1.2 points for Cosmos Policy and 1.8 for $\pi_{0.5}$ at $1.8\times$ the step limit, and detecting the failure is not enough either: with a fixed retry, in which the arm retreats and the embodied model tries again, success stays at the level of the embodied model even with the oracle as detector, consistent with Section~\ref{sec:eval}. Spotter improves both models: with the GPT judge and one learned example it reaches 73.5 and 71.8, 5.6 and 7.5 points above the embodied model under the same budget, mainly on pick-and-place (10.3 and 9.0 points, against 3.3 and 6.7 on the other tasks), where a missed grasp is common and can be retried. Harness VLA with Qwen, which consults the VLM at every round and is given 5,000 steps, reaches 71.3 and 69.1 when its VLM can also see the task's success predicate, as in its released code, but without it falls by 6.1 and 8.6 points, most sharply on the tasks that operate a mechanism (9.2 and 11.7 points), to 65.2 and 60.5, below the embodied model alone. A VLM-led design relies on the language model to drive every step: the original work uses the Codex and Claude Code agents as planners, and a weaker model such as Qwen works only with privileged information. Spotter needs neither: with the same Qwen model and no privileged information it gains 3.8 and 4.1 points, and with one learned example it exceeds privileged Harness VLA on both models with either judge (71.8 and 70.4 with Qwen, 73.5 and 71.8 with GPT-6 Astra). Removing the screener, so that the judge examines every supervision window, adds about two points (70.0 to 72.0 and 68.8 to 70.8), the price of keeping the frontier VLM off the common path. On RoboTwin 2.0, with longer bimanual tasks, the gap between the two judges widens (Figure~\ref{fig:results}(b)): from $\pi_{0.5}$ at 47.2\%, Spotter reaches 51.2\% with Qwen and 57.0\% with GPT-6 Astra, a 5.8-point gap against 1.7 and 1.4 on RoboCasa.

\subsection{Real-Robot Validation}
\label{sec:real}

\begin{figure}[!htbp]
\begin{center}
\includegraphics[width=\textwidth]{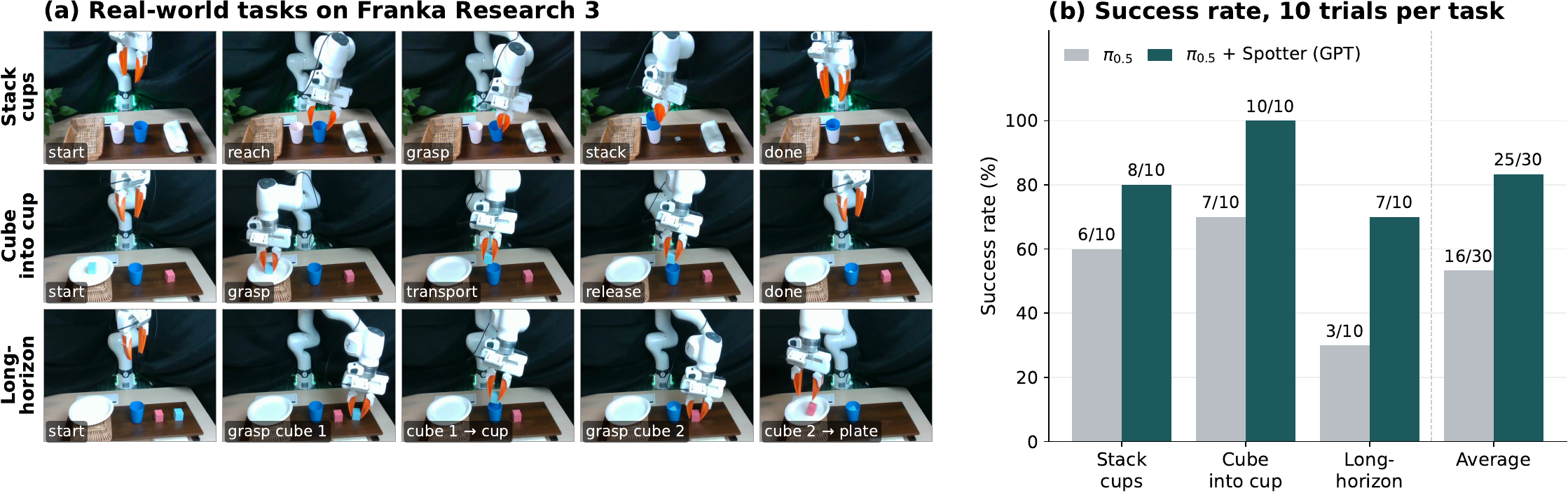}
\end{center}
\caption{Real-robot validation on a Franka Research 3. (a) The three tasks (Section~\ref{sec:real}). (b) Success rate over 10 trials per task for $\pi_{0.5}$ alone and with Spotter (GPT), 1-shot.}
\label{fig:real}
\end{figure}

We fine-tune $\pi_{0.5}$ on teleoperated demonstrations for three tasks on a Franka Research 3: stacking cups, placing a cube into a cup, and a long-horizon task that chains two subtasks (one cube into the cup, then another onto the plate). Adding Spotter (GPT) raises average success from 53\% (16/30) to 83\% (25/30), and the largest gain is on the long-horizon task (30\% $\rightarrow$ 70\%), where a single mistake in either subtask is enough to fail the task without intervention (Figure~\ref{fig:real}).

\subsection{Cost of Supervision}
\label{sec:cost}

\begin{figure}[!htbp]
\begin{center}
\includegraphics[width=\linewidth]{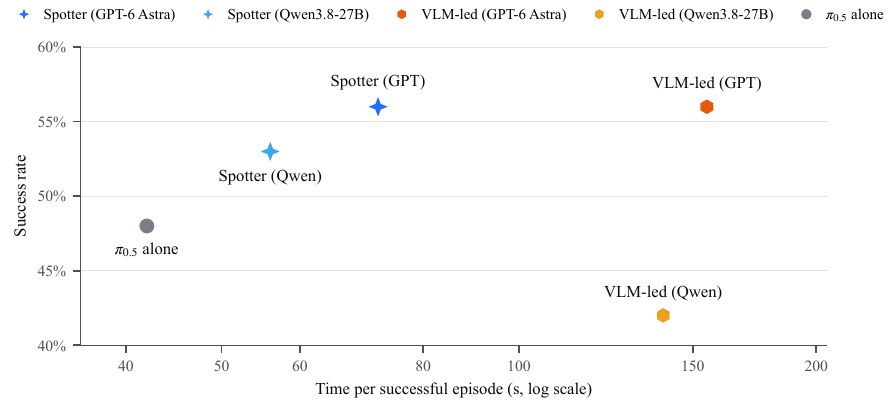}
\end{center}
\caption{Success rate against time per successful episode on RoboTwin 2.0 (Hard) with $\pi_{0.5}$, in a supplementary run of 150 episodes (50 tasks, 3 episodes each); up and left is better. Every method has $1.8\times$ the official step limit and no privileged information, and Spotter uses one learned example. Qwen times assume one H200 serving one request at a time.}
\label{fig:robotwin-time}
\end{figure}

Figure~\ref{fig:robotwin-time} compares success rate and time on RoboTwin 2.0 with both VLMs, with every method, Harness VLA included, at $1.8\times$ the step limit. Spotter matches or exceeds the VLM-led design with either VLM while staying close to the policy alone in time. With GPT-6 Astra both reach 56\%, but a successful episode takes 72\,s with Spotter against 155\,s when the VLM leads. With Qwen the VLM-led design falls to 42\%, below the policy alone at 48\%, whereas Spotter reaches 53\% in 56\,s, against 42\,s for the policy alone.

\begin{figure}[!htbp]
\begin{center}
\includegraphics[width=\linewidth]{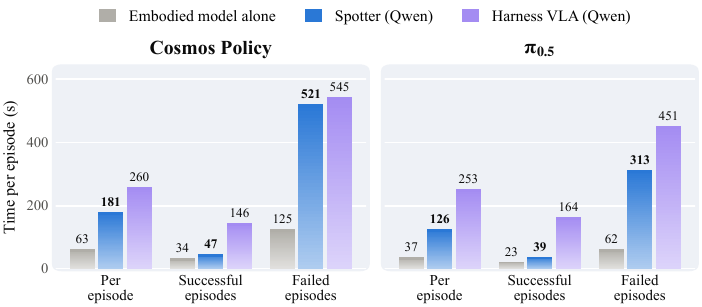}
\end{center}
\caption{Wall-clock time per episode on RoboCasa, overall and on successful and failed episodes: the embodied model alone at $1.8\times$ the step limit, Spotter (Qwen, full context), and Harness VLA (Qwen, with privileged info), whose supervision blocks execution at every round. Time and tokens for every configuration are in Table~\ref{tab:cost}.}
\label{fig:time}
\end{figure}

On RoboCasa, Figure~\ref{fig:time} compares the time per episode with the embodied model alone and with Harness VLA; Tables~\ref{tab:cost} and~\ref{tab:calls-full} in Appendix~\ref{app:setup} report time and tokens for every configuration and the judge calls and repairs behind them. On successful episodes Spotter stays close to the embodied model alone: with Qwen, the judge is consulted less than once per episode and almost never repairs, so such an episode takes only 13\,s longer on Cosmos Policy and 16\,s on $\pi_{0.5}$, 3\% and 6\% of the extra time of a failed episode, which takes $4.2\times$ and $5.0\times$ as long as without Spotter. The small local screener does much of the work: execution continues while chunks are checked, hiding 36 to 38\,s of supervision per episode on Cosmos Policy and 29 to 30\,s on $\pi_{0.5}$ (about 60\% and 80\% of the policy's own execution time), and with GPT-6 Astra it cuts judge calls per episode by 37\% and tokens by 66\% at a cost of 2.0 points of success rate. Section~\ref{sec:analysis} analyzes the cost of the single demonstration.

\section{Analysis}
\label{sec:analysis}

\textbf{Why VLM-led supervision is slow where it is not needed.} The gap to Harness VLA is largest where supervision matters least: on successful Cosmos Policy episodes, Harness VLA with Qwen takes 146\,s and 46k tokens, against 47 to 62\,s and 9k to 33k for Spotter (Qwen) and 34\,s for the policy alone (Table~\ref{tab:cost}). The embodied model could have finished these episodes alone, but a VLM-led planner must fix how long each call runs before seeing how it turns out, so it stops to check even when nothing is wrong, spending so many steps that under the $1.8\times$ limit its success rate falls to about 50\%, with 1,802 steps per episode on average (Appendix~\ref{app:setup}). Checking less means running the embodied model longer between calls, so an error early in a call goes unnoticed until the call returns, by which time the robot may have gone far down a failure. Spotter avoids this choice: every chunk is included in a screening check, with the supervisor trailing execution by at most $K$ chunks (Eq.~(\ref{eq:lead})), while frontier VLM calls are reserved for flagged windows and repair verification, keeping successful episodes only 13 to 28\,s longer than with the policy alone.

\begin{figure}[!htbp]
\begin{center}
\includegraphics[width=0.62\linewidth]{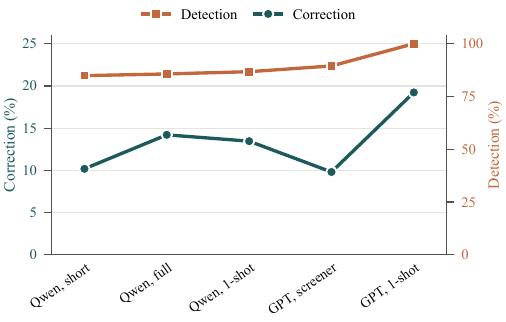}
\end{center}
\caption{The judge on RoboCasa, with Cosmos Policy and $\pi_{0.5}$ pooled. Detection: share of the episodes it intervened in that the embodied model alone fails on the same scene. Correction: share of these errors corrected by a single repair, not by three attempts as in Section~\ref{sec:eval}.}
\label{fig:correct}
\end{figure}

\textbf{How often does the judge correct an error?} Section~\ref{sec:eval} found that, even once an oracle has caught the error, the embodied model corrects fewer than one failure in ten on its own. We ask the same of Spotter, which finds the error itself and repairs it with its own primitive plan. Let $\mathcal{I}$ be the episodes in which the judge intervened, $\mathcal{E}\subseteq\mathcal{I}$ those the embodied model alone fails on the same scene, and $\mathcal{C}\subseteq\mathcal{E}$ those in which a single repair makes the episode succeed. Figure~\ref{fig:correct} reports
\begin{equation}
\mathrm{Detection} = \frac{|\mathcal{E}|}{|\mathcal{I}|}, \qquad \mathrm{Correction} = \frac{|\mathcal{C}|}{|\mathcal{E}|}.
\end{equation} The judge intervenes where it should: with Qwen as both screener and judge, about 85\% of the episodes it intervenes in are ones the embodied model alone fails, and GPT-6 Astra as the judge behind the Qwen screener raises this further. A single repair corrects about 10\% to 20\% of these errors, against 7.5\% to 9.7\% for the embodied model with three attempts and an oracle that had already found the error (Section~\ref{sec:eval}); with one learned example each, GPT corrects 19.2\% against 13.5\% for Qwen.

\textbf{Why the single demonstration does not always show up in the average.} On $\pi_{0.5}$ the demonstration helps uniformly, by exactly 2.0 points in all three columns. On Cosmos Policy the aggregate barely moves, from 71.7 to 71.8, but this hides a redistribution: the demonstration is worth $+4.2$ points on the eight pick-and-place tasks and $-1.9$ on the sixteen others, and since the latter carry two thirds of the weight, the two nearly cancel. It also costs $2.5\times$ the tokens (Table~\ref{tab:cost}). For Cosmos Policy the demonstration therefore buys a different distribution of successes rather than more of them; for $\pi_{0.5}$ it simply buys more. The single demonstration does not lengthen episodes despite $2.5\times$ and $2.2\times$ the tokens: time per episode changes by $-4\%$ and $0\%$, and the judge is called 7\% less often per episode on both models, mainly because the success rate rises by 2.0 points on $\pi_{0.5}$ and failed episodes need 15\% fewer calls on Cosmos Policy.

\section{Conclusion}
\label{sec:conclusion}

Recovering from a failure takes two abilities, producing a correction and recognizing the right one, and embodied models learned only from success have neither: they rarely repair a failure even when it is caught for them, and the scorers built on them lose their skill once the policy has failed. What is missing is not more samples from the embodied model but a different reader of the episode: a general vision-language model that reads a longer stretch of it with telemetry and past attempts, adjusts from the failed attempt, and checks whether its repair worked. This reflection need not take control away from the embodied model: Spotter adds continuous, lightweight attention between the two, keeping the embodied model running while a cheap screener watches and calling the general model for flagged windows and repair verification, so it helps most where one mistake would end the episode and costs little elsewhere. Its lag is undone by returning the arm rather than resetting the scene, so the approach carries over to a real robot. Errors that ruin the scene the first time they occur remain out of reach; anticipating them, perhaps with a world model, is the next step toward embodied models that recover from their own mistakes.

\subsubsection*{Acknowledgments}

We thank MainCode for funding this work.

\bibliography{iclr2027_conference}
\bibliographystyle{iclr2027_conference}

\appendix
\section{Implementation Details of the Error-Correction Evaluation}
\label{app:eval}

This appendix gives the details of the two experiments in Section~\ref{sec:eval}. Both use the official checkpoints of Cosmos Policy (2B, fine-tuned on RoboCasa) and $\pi_{0.5}$, run closed-loop in RoboCasa. All privileged quantities below (object poses, gripper--object distance, grasp state, and the task success predicate) are read from the simulator by the evaluation harness only. They are never shown to the embodied model or to the Spotter supervisor; the evaluation harness uses the success predicate only for scoring and episode termination.

\subsection{Producing a correction}

\textbf{Failure detection.} After every chunk the oracle checks whether a grasp was attempted and failed. A grasp counts as attempted when the gripper has come within 5\,cm of the target object and its aperture has changed by at least 0.008 within the chunk. It counts as failed when the object is not held at the end of the chunk, or was held at the midpoint of the chunk and has been dropped by its end.

\textbf{Resampling.} The simulator state is saved at every chunk boundary. When a failure is detected, the state at the start of the failed chunk is restored and the model is queried again from that state with a new random seed; the resulting rollout is executed to the end of the episode under the original step budget. This is repeated for three seeds, and all three rollouts are executed. The failure counts as corrected if the object is lifted in any of the three.

\textbf{Language feedback.} When a failure is detected, execution stops without rewinding. A VLM is shown the current camera images and the task instruction, is told that the grasp has failed, and is asked for the cause of the failure and for a short imperative description of what to do differently. This text is given to the embodied model through its language input, and execution resumes. If the next grasp attempt also fails, the procedure repeats, for at most three rounds; success in any round counts as corrected. We run this protocol with $\pi_{0.5}$, whose language input accepts free-form text.

\textbf{Metric.} Let $\mathcal{F}$ be the set of oracle-detected failures, with $|\mathcal{F}|=320$ in every setting, drawn from 1{,}200 episodes per model, and let $s_k(f)\in\{0,1\}$ indicate that attempt $k$ after failure $f$ lifts the object. The correction rate is
\begin{equation}
\mathrm{CR} = \frac{1}{|\mathcal{F}|}\sum_{f\in\mathcal{F}} \max_{k\le 3} s_k(f).
\end{equation}

\subsection{Judging a correction}

\textbf{Saved states and candidates.} While the policy runs, the harness evaluates a trigger condition at every chunk boundary. When it fires, the full simulator state is saved. From each saved state we restore the simulator and sample eight candidates from the same policy and checkpoint with different random seeds. Each candidate is executed in the real simulator for 32 steps: the 16 steps of the sampled chunk, followed by 16 steps of the continuation of the same plan. A candidate is labeled by the benchmark's official grasp predicate at step 32, and a success observed at any earlier step is kept.

\textbf{Triggers.} In the \emph{post-failure} regime the trigger is the failure condition defined above: the policy attempted a grasp in the chunk that just ended and did not succeed, so the state is saved at the instant of the failure. In the \emph{normal} regime the trigger is the first chunk of an episode in which the target becomes grasped, and the state is saved at the start of that chunk, that is, just before the successful grasp; it fires at most once per episode. The saved state therefore lies on an ordinary successful trajectory. The two regimes share the candidate count, the horizon, the labeling rule, and the scorers, and differ only in the trigger.

\textbf{Metric.} Only states whose eight candidates contain both a success and a failure can be used, since ranking is undefined otherwise. For these states we form the set $\mathcal{P}$ of every (successful, failed) pair of candidates $(c^+,c^-)$ from the same state and report, for a scorer $r$, the fraction of pairs in which it scores the successful candidate higher, with ties counted as one half:
\begin{equation}
\mathrm{Acc}(r) = \frac{1}{|\mathcal{P}|}\sum_{(c^+,c^-)\in\mathcal{P}} \Big(\mathbf{1}\big[r(c^+)>r(c^-)\big] + \tfrac{1}{2}\,\mathbf{1}\big[r(c^+)=r(c^-)\big]\Big).
\end{equation} Confidence intervals are 95\% intervals from 1{,}000 bootstrap resamples over states. The post-failure regime yields 991 usable states and 10{,}974 pairs, and the normal regime 432 states and 4{,}853 pairs. Usable states are rarer in the normal regime because candidates drawn from a state that is about to complete a grasp almost always succeed together.

\textbf{Scorers.} Every scorer assigns one number to a candidate at the saved state, before the candidate is executed, using only what is available at deployment, with one exception noted below. \emph{Random} draws a uniform score and verifies that the pipeline is unbiased. \emph{Cosmos value} is the output of the policy's own value head for the candidate. \emph{Consistency-Consensus} is the agreement between the candidate's imagined future and the mean imagined future of the eight candidates. \emph{GeoBoN} is the cross-view depth reprojection consistency between the imagined third-person and wrist frames, computed with a frozen geometry model. \emph{RCS} re-noises the imagined future at a noise level from the training distribution and measures how closely the same model denoises it back. \emph{Consistency-Exploring} compares the imagined future with the frames actually observed after the candidate is executed. It needs the real outcome and cannot be deployed; we include it as a diagnostic upper bound on what Consistency-Consensus approximates, which our design provides at no extra cost because every candidate is executed. \emph{VLM judge} is the judge of Spotter, Qwen3.8-27B or GPT-6 Astra, asked for each candidate separately how likely it is to grasp the object, from 0 to 1. It sees only images, those of the chunk leading to the saved state and the candidate's imagined future frames, with no earlier history of the episode, no telemetry, and no retrieved lessons or demonstration.

\section{Detailed Experimental Setup}
\label{app:setup}

\textbf{Tasks and seeds.} On RoboCasa we use the 24 official tasks. The eight pick-and-place tasks are PnPCounterToCab, PnPCabToCounter, PnPCounterToSink, PnPSinkToCounter, PnPCounterToMicrowave, PnPMicrowaveToCounter, PnPCounterToStove and PnPStoveToCounter. The sixteen others operate doors (OpenSingleDoor, CloseSingleDoor, OpenDoubleDoor, CloseDoubleDoor), drawers (OpenDrawer, CloseDrawer), the stove (TurnOnStove, TurnOffStove), the sink (TurnOnSinkFaucet, TurnOffSinkFaucet, TurnSinkSpout), the coffee machine (CoffeeSetupMug, CoffeeServeMug, CoffeePressButton) and the microwave (TurnOnMicrowave, TurnOffMicrowave). Formal RoboCasa evaluation uses 1{,}200 episodes: base seed 500 with episode indices 0--49 of each task and object instances from the held-out split. The released code also retains a fixed 96-episode subset (four episodes per task at seed 500) for implementation debugging only; it does not replace the formal evaluation set. The environment seed is derived as $256(s+e)$ from the base seed $s$ and episode index $e$. Example episodes are mined from separate learning runs with base seed 195; the example library is read-only during evaluation. On RoboTwin 2.0, which is simulated, we test on the official 50 tasks in the Hard setting, with seed 0 and the first 10 episodes of each task; $\pi_{0.5}$ is trained on clean data only.

\textbf{Step budget.} The official RoboCasa limit is 500 steps for most tasks, 300 for CoffeePressButton, 600 for PnPCounterToMicrowave, CoffeeSetupMug and CoffeeServeMug, 700 for PnPCounterToSink and CloseDoubleDoor, and 1{,}000 for OpenDoubleDoor. On RoboTwin 2.0 the official limits range from 400 to 1{,}700 steps. The main RoboCasa comparison applies the same $1.8\times$ multiplier to the official limit for the baseline and Spotter. After initial scene stabilization, every executed environment step counts toward this shared budget, including repair motions, gripper actions and controller calibration.

\textbf{Chunks and screening.} The default is bounded asynchronous supervision. Cosmos Policy executes 16 steps per inference, one chunk, and the screener checks every two chunks, or 32 steps. $\pi_{0.5}$ outputs 50 steps and executes the first 25, one chunk, and the screener checks every chunk. The lead limit is $K=4$ chunks on Cosmos Policy and $K=2$ on $\pi_{0.5}$, so the supervisor trails execution by at most 64 and 50 policy steps. Delayed checks combine the images and telemetry of completed, unchecked windows, including windows accumulated during judge deliberation. Policy chunks executed within a repair are also screened before the next repair decision. The screener flags a check if its risk score, from 0 to 10, is at least 3, if screening fails or returns an invalid score, or if the gripper closes below 12\,mm in any chunk (including a fully closed, 0\,mm empty grasp; the threshold is 4\,mm on the mug tasks, where holding a handle leaves 6 to 12\,mm). The judge is called after two consecutive flags, or at once on newly screened empty-gripper evidence if it has not intervened in the previous two checks. The next supervised window after a repair is checked by the judge even when the screener does not flag it; this check cannot trigger another repair. A repair has at most 8 primitives and 3 plan segments. Table~\ref{tab:context} lists what differs between the three configurations.

\textbf{Inputs to the supervisor.} The supervisor receives time-ordered montages from the left and right third-person cameras and the wrist camera, with pending windows combined when supervision is delayed. Repair decisions also receive a fresh checkpoint image. The text includes the instruction, action history, remaining budget, proprioceptive telemetry, empty-grasp events, retrieved general lessons, the primitive interface and the expectation from the previous repair. With one learned example, the prompt adds one example episode of at most 12 images. No object pose or success signal is given.

\textbf{Serving.} Qwen3.8-27B-FP8 runs on vLLM with prefix caching, one replica on each of four NVIDIA H20 GPUs, with a context of 131,072 tokens and up to 60 images per request (32,768 tokens and 20 images in the short-context runs). The judge uses temperature 0.2 and at most 3,000 output tokens, the screener temperature 0 and at most 120, both with thinking disabled. GPT-6 Astra is called through the Responses API with medium reasoning effort, and the conversation continues on the server side. In Spotter (GPT) the screener is still the local Qwen.

\textbf{General lessons.} The simulator's reset is used only while learning: on training data or on seeds other than the test seeds, we let GPT reset and retry, and summarize what it finds into a library of general lessons. General means physical knowledge that holds across tasks, for example that a long object such as a cola bottle should be grasped in the middle rather than at an end, where it tends to slip. Privileged or task-specific information, such as how many degrees to rotate in a particular scene, is removed.

\textbf{Learning.} The released general-lesson library contains 77 entries and is frozen during evaluation. A single demonstration, when enabled, is retrieved from a separately prepared example library. Selection prefers the same policy and task, then the same task, then a similar task, then the whole library, preferring a repair example without overriding task relevance.

\textbf{Timing.} Table~\ref{tab:cost} excludes scene construction. Times are omitted for GPT-6 Astra because they include network latency. Qwen token counts include prefix-cache hits. The $\pi_{0.5}$ short-context configuration screens every two chunks, or 50 steps.

\begin{table}[!htbp]
\caption{Time and tokens per episode, overall and on successful and failed episodes. \emph{Async}: wall-clock time with parallel supervision; \emph{Sync}: time if supervision blocked execution, as in Harness VLA (with privileged info) at every round. Cosmos Policy and $\pi_{0.5}$: the models alone at $1.8\times$ the step limit. Details in Appendix~\ref{app:setup}.}
\label{tab:cost}
\begin{center}
\small
\setlength{\tabcolsep}{3pt}
\begin{tabular*}{\linewidth}{@{\extracolsep{\fill}}lrrrrrrrrr@{}}
\toprule
 & \multicolumn{3}{c}{\textbf{Per episode}} & \multicolumn{3}{c}{\textbf{Successful episodes}} & \multicolumn{3}{c}{\textbf{Failed episodes}} \\
\cmidrule(lr){2-4} \cmidrule(lr){5-7} \cmidrule(lr){8-10}
\textbf{Method} & Async & Sync & Tokens & Async & Sync & Tokens & Async & Sync & Tokens \\
 & (s) & (s) & (k) & (s) & (s) & (k) & (s) & (s) & (k) \\
\midrule
Cosmos Policy & 63 & -- & -- & 34 & -- & -- & 125 & -- & -- \\
Spotter (Qwen) & & & & & & & & & \\
\quad -short context & 158 & 196 & 82  & 54 & 77 & 18  & 395 & 467 & 227 \\
\quad -full context  & 181 & 217 & 87  & 47 & 71 & 9   & 521 & 587 & 283 \\
\quad -1-shot        & 173 & 210 & 218 & 62 & 86 & 33  & 454 & 524 & 689 \\
Spotter (GPT) & & & & & & & & & \\
\quad -with screener & -- & -- & 156 & -- & -- & 48  & -- & -- & 411 \\
\quad -no screener   & -- & -- & 456 & -- & -- & 232 & -- & -- & 1030 \\
\quad -1-shot        & -- & -- & 226 & -- & -- & 54  & -- & -- & 705 \\
Harness VLA (Qwen) & -- & 260 & 135 & -- & 146 & 46 & -- & 545 & 356 \\
\midrule
$\pi_{0.5}$ & 37 & -- & -- & 23 & -- & -- & 62 & -- & -- \\
Spotter (Qwen) & & & & & & & & & \\
\quad -short context & 70  & 99  & 43  & 27 & 43 & 9   & 158 & 212 & 113 \\
\quad -full context  & 126 & 156 & 77  & 39 & 58 & 11  & 313 & 370 & 220 \\
\quad -1-shot        & 126 & 156 & 170 & 41 & 61 & 26  & 328 & 382 & 514 \\
Harness VLA (Qwen) & -- & 253 & 170 & -- & 164 & 84 & -- & 451 & 362 \\
\bottomrule
\end{tabular*}
\end{center}
\end{table}

\begin{table}[!htbp]
\caption{Judge calls and executed repairs per episode for every configuration, on successful and failed episodes. The $\pi_{0.5}$ short-context row screens every two chunks, or 50 steps. For Harness VLA every interaction round counts as a repair.}
\label{tab:calls-full}
\begin{center}
\begin{tabular*}{\linewidth}{@{\extracolsep{\fill}}lrrrr@{}}
\toprule
 & \multicolumn{2}{c}{\textbf{Judge calls}} & \multicolumn{2}{c}{\textbf{Repairs}} \\
\cmidrule(lr){2-3} \cmidrule(lr){4-5}
\textbf{Method} & Succ. & Fail. & Succ. & Fail. \\
\midrule
Cosmos Policy + Spotter (Qwen) & & & & \\
\quad -short context & 1.1 & 10.9 & 0.0 & 1.5 \\
\quad -full context  & 0.6 & 10.9 & 0.0 & 3.7 \\
\quad -1-shot        & 0.9 & 9.3  & 0.1 & 3.1 \\
Cosmos Policy + Spotter (GPT) & & & & \\
\quad -with screener & 2.5 & 13.1 & 0.2 & 4.0 \\
\quad -no screener   & 5.5 & 17.8 & 0.1 & 2.8 \\
\quad -1-shot        & 1.4 & 14.2 & 0.2 & 3.2 \\
Cosmos Policy + Harness VLA (Qwen) & -- & -- & 7.7 & 39.6 \\
\midrule
$\pi_{0.5}$ + Spotter (Qwen) & & & & \\
\quad -short context & 0.6 & 5.4  & 0.1 & 0.8 \\
\quad -full context  & 0.6 & 9.9  & 0.0 & 2.4 \\
\quad -1-shot        & 0.7 & 9.4  & 0.1 & 2.6 \\
\bottomrule
\end{tabular*}
\end{center}
\end{table}

\textbf{Compared methods.} We also compare with other embodied models in Figure~\ref{fig:results}. On RoboCasa, RLDX-1 \citep{kim2026rldx} is taken from its technical report, and Cosmos Policy, GR00T-N1.5 (the N1.5 release of \citealp{bjorck2025gr00t}), UWM \citep{ZhuC-RSS-25}, $\pi_0$ \citep{BlackK-RSS-25}, Video Policy \citep{liang2025video}, and FLARE \citep{pmlr-v305-zheng25a} from the Cosmos Policy paper, all under the official step limit. On RoboTwin 2.0, GR00T-N1.7 (the N1.7 release of \citealp{bjorck2025gr00t}) and LingBot-VLA \citep{wu2026pragmatic} are taken from Harness VLA \citep{zhang2026harness}, which evaluates 50 tasks with 5 seeds each, and X-WAM \citep{guo2026unified} is run by us. In the figure, Ours-Qwen and Ours-GPT are Spotter (Qwen) and Spotter (GPT) with one learned example under $1.8\times$ the step limit, on Cosmos Policy for RoboCasa and on $\pi_{0.5}$ for RoboTwin 2.0.

\textbf{Harness VLA.} To compare running cost, we also run Harness VLA with Qwen as its VLM. We keep its original settings but cut the maximum number of interaction rounds from 100 to 40 and raise its step budget to 5,000. Under the same step limit as the other methods its success rate falls to about 50\%: its rounds spend many steps while the task is still under way, so the limit is reached halfway through. It uses 1,802 steps per episode on average, more than the $1.8\times$ limit of most tasks. In the supplementary RoboTwin 2.0 comparison (Figure~\ref{fig:robotwin-time}), run with both Qwen and GPT-6 Astra as its VLM, Harness VLA instead has $1.8\times$ the official step limit, the same budget as Spotter.

\section{Primitive Library}
\label{app:primitives}

Table~\ref{tab:primitives} lists the primitives from which the judge composes a repair plan. In the RoboCasa executor, a move target is either an explicit world-coordinate vector $(x,y,z)$ or one pixel $(r,c)$ in a selected camera view, with an optional vertical offset. For a pixel target, the executor reads the calibrated depth $d=D_t[r,c]$ and applies the inverse camera projection to $(cd,rd,d,1)$, then adds the vertical offset. It uses that single depth sample, not a median over a bounding box; depth remains internal to the executor and is not shown to the VLM.

\end{document}